\documentclass{article}
\usepackage{ijcai26}

\usepackage{times}
\usepackage{soul}
\usepackage{url}
\usepackage[hidelinks]{hyperref}
\usepackage[utf8]{inputenc}
\usepackage[small]{caption}
\usepackage{graphicx}
\usepackage{amsmath}
\usepackage{amsthm}
\usepackage{booktabs}
\usepackage{algorithm}
\usepackage{algorithmic}
\usepackage[switch]{lineno}
\DeclareMathOperator*{\argmax}{arg\,max}

\title{Controlling Decision Drift in Multimodal Sentiment Analysis \\with Missing Modalities}

\author{
	Chenglizhao Chen$^{1,2}$
	\and
	Yuchen Cao$^{1,2}$
	\and
	Xinyu Liu$^{1,2}$\thanks{Corresponding author.}
	\and
	Mengke Song$^{1,2}$
	\and
	\\Guisheng Zhang$^{1,2}$
	\And
	Xiaomin Yu$^3$\\
	\affiliations
	$^1$Qingdao Institute of Software, College of Computer Science and Technology, \\China University of Petroleum (East China)\\
	$^2$Shandong Key Laboratory of Intelligent Oil \& Gas Industrial Software\\
	$^3$The Hong Kong University of Science and Technology (Guangzhou)\\
	\emails
	\{cclz123, cyc021109, liuxy001005, songsook, sdut\_guisheng\}@163.com,
	yuxm02@gmail.com
}

\begin{document}
	
	\maketitle
	
	\begin{abstract}
		Multimodal sentiment analysis relies on textual, acoustic, and visual signals, yet real-world data often suffer from modality missing and quality imbalance. Existing methods generate features for modality missing from available ones, but differences in expression mechanisms and sentiment dynamics across modalities may cause the generated features to deviate from true distributions and mislead prediction. In addition, unreliable modalities may dominate fusion, resulting in representation shift across modality combinations and unstable sentiment representations. To address these challenges, we propose a two-level reference alignment framework. The framework introduces stable references at the feature representation and sentiment decision levels to improve robustness under modality missing. First-level reference alignment leverages complete-modality samples to constrain representations and align different modality combinations into a shared sentiment space. Second-level reference alignment enforces cross-modal consistency at the decision level by suppressing unreliable modalities through prototype retrieval and voting. As a result, the framework maintains stable and reliable sentiment predictions under diverse missing-modality patterns. Experiments on CMU-MOSI and CMU-MOSEI show consistent improvements across various missing-modality settings. Under full-modality input, the proposed method achieves state-of-the-art performance, with ACC of 86.28\% and 85.88\%, and F1 of 86.24\% and 85.86\%.
	\end{abstract}
	
	\begin{figure}[!t]
		\centering{\includegraphics[width=1\linewidth]{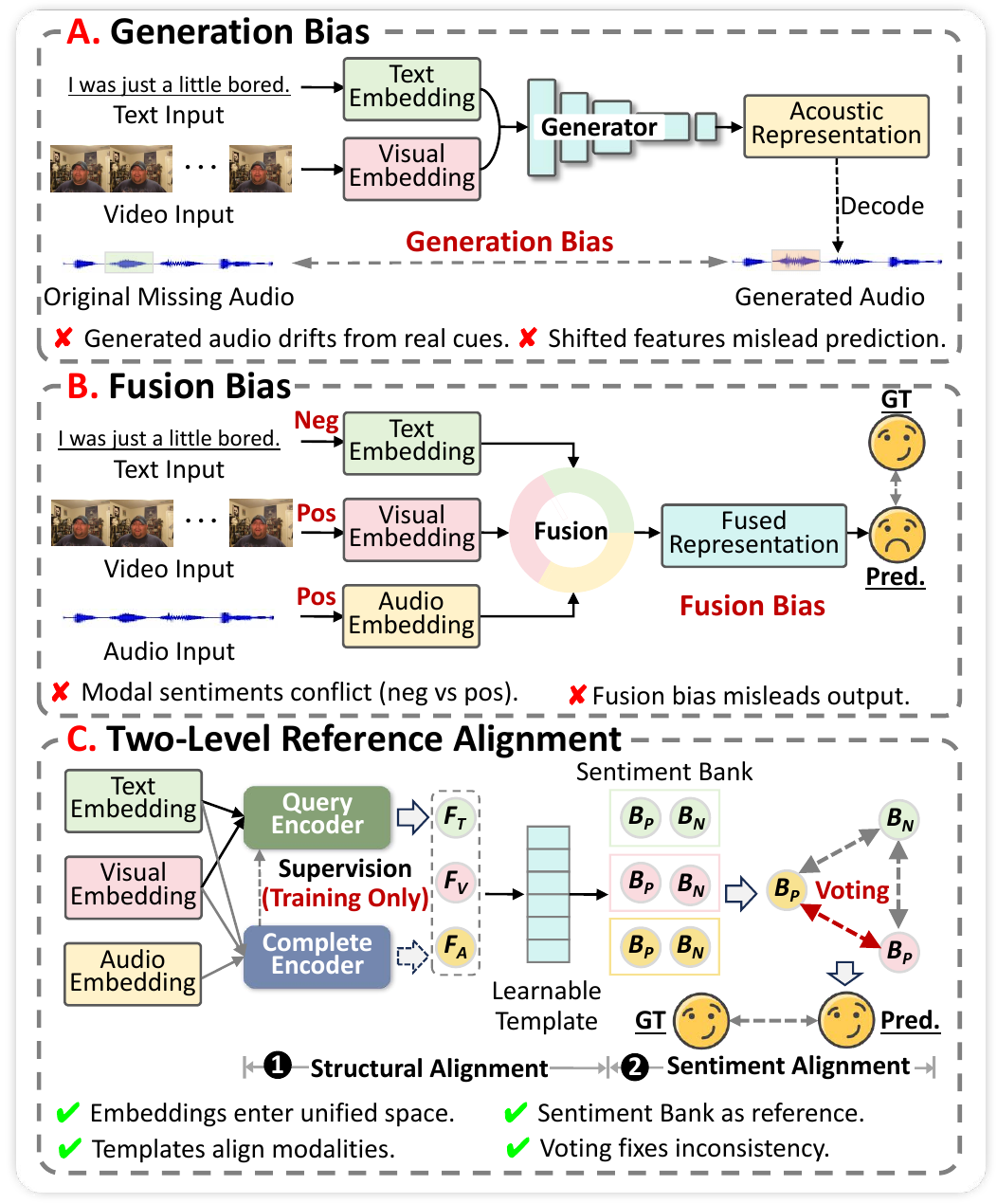}}
		\caption{Illustration of generation bias, fusion bias, and Two-Level Reference Alignment. (A) Missing modalities shift generated features away from sentiment cues. (B) Unreliable modalities dominate fusion and mislead prediction. (C) Feature- and decision-level references align representations and reduce both biases.}
		\label{fig:motivation}
	\end{figure}
	
	\section{Introduction}
	
	The objective of sentiment analysis is to identify human sentiments during communication, which provides a foundation for understanding behaviors and social interactions~\cite{liu2022sentiment}. Sentiments are expressed through textual content, vocal intonation, and body language, making multimodal sentiment analysis necessary for capturing emotional dynamics more comprehensively~\cite{poria2017review}. Integrating textual, acoustic, and visual information enables richer affective representation and supports broader human-computer interaction applications~\cite{zadeh2018multi}.
	
	Specifically, in real-world deployments, multimodal observations are often missing or unreliable due to environmental conditions, occlusions, and system failures~\cite{wu2024deep,reza2024robust}. Such unreliability may arise from background speech, motion blur, illumination variation, occlusion, or sensor failure, which commonly lead to unstable representations and biased fusion decisions. These situations introduce inconsistent cues and make it difficult to obtain accurate sentiment representations~\cite{yu2021learning}. Under modality missing scenarios, many methods generate missing-modality features from available modalities to preserve the multimodal prediction structure~\cite{guo2024multimodal}. However, due to differences in expression mechanisms and sentiment dynamics, generated features may deviate from the true feature space and incur generation bias, as illustrated in Figure~\ref{fig:motivation}(A)~\cite{dai2025unbiased}, further misleading fusion and drifting predictions toward incorrect semantic regions~\cite{liu2024modality}.
	
	Even when all modalities are available, multimodal systems still face quality imbalance. Modalities affected by noise, occlusion, or abnormal expressions may be treated as reliable signals during fusion~\cite{arevalo2017gated,mao2023robust}. Although such corruptions differ at the input level, they commonly reduce modality reliability and introduce inconsistent sentiment evidence. As shown in Figure~\ref{fig:motivation}(B), unreliable sources may be over-weighted in the fused representation, pushing predictions away from the correct sentiment target and causing fusion bias~\cite{baltruvsaitis2018multimodal}.
	
	To address generation bias and fusion bias, we introduce a two-level reference alignment (TLRA) framework that provides controllable cross-modal references at both the feature and sentiment decision levels, as illustrated in Figure~\ref{fig:motivation}(C). TLRA uses complete-modality information as stable references to guide feature alignment under modality missing, thereby reducing the deviation of generated or completed features from the true sentiment space. It further models cross-modal consistency at the decision level through reference-driven aggregation, suppressing the influence of unreliable modalities on final prediction. Rather than designing noise-specific correction modules, TLRA treats missing and unreliable modalities as a unified reliability degradation problem and constrains their effects through reference alignment.
	
	The contributions of this work are:
	\begin{itemize}
		\item
		A TLRA framework that introduces cross-modal references at both feature and sentiment decision levels to address generation bias under modality missing and fusion bias caused by unreliable modalities.
		\item
		A reference-constrained feature alignment strategy that leverages complete-modality samples as stable semantic anchors to reduce generation bias and stabilize representations under modality missing.
		\item
		A reference-driven decision alignment strategy that models sentiment-level cross-modal consistency with prototype-guided references and reduces the influence of potentially unreliable modalities during fusion.
		\item
		Extensive experiments on CMU-MOSI and CMU-MOSEI demonstrate that the proposed method outperforms representative methods under various missing-modality scenarios and achieves state-of-the-art performance under complete modalities.~\footnote{https://github.com/LiuXY3366/TLRA}
	\end{itemize}

	\section{Related Works}
	
	\subsection{Sentiment Analysis}
	Early multimodal sentiment analysis methods commonly assume complete and well-aligned modalities, improving affective modeling through cross-modal attention, feature interaction, and temporal fusion~\cite{liu2018efficient,tsai2019multimodal,yu2024spikemo}. However, real-world modalities can be degraded by noise or missing values, making these methods sensitive to modality stability~\cite{baltruvsaitis2018multimodal,zeng2022mitigating,li2024toward}. Existing studies address this issue using cross-modal generation, text-dominant recovery, or robustness training~\cite{neverova2015moddrop,pham2019found,lin2023missmodal}. Nevertheless, because modalities differ in expression mechanisms and temporal dynamics, recovered features may not align with the true sentiment structure and can produce unstable or conflicting fused representations~\cite{lin2023missmodal,zeng2022mitigating}. Unlike generation-based recovery, our method introduces stable cross-modal references to constrain sentiment representations under missing or quality-imbalanced modality conditions.
	
	\subsection{Multimodal Fusion}
	In multimodal sentiment analysis, fusion performance is strongly affected by modality quality, expression stability, and signal conditions~\cite{liu2022sentiment}. Prior studies improve robust fusion by estimating modality confidence, adaptively weighting modality contributions, or imposing consistency objectives to reduce uncertainty-induced drift~\cite{arevalo2017gated,zadeh2018memory,wang2019words,xie2024trustworthy,han2021improving,hazarika2020misa,yu2026anisotropic,yu2026modality}. In contrast, this work introduces prototype-based decision references to align modality-wise predictions and enhance cross-modal consistency when modality quality varies in practical scenarios.

	\begin{figure*}[!t]
		\centering{\includegraphics[width=1\linewidth]{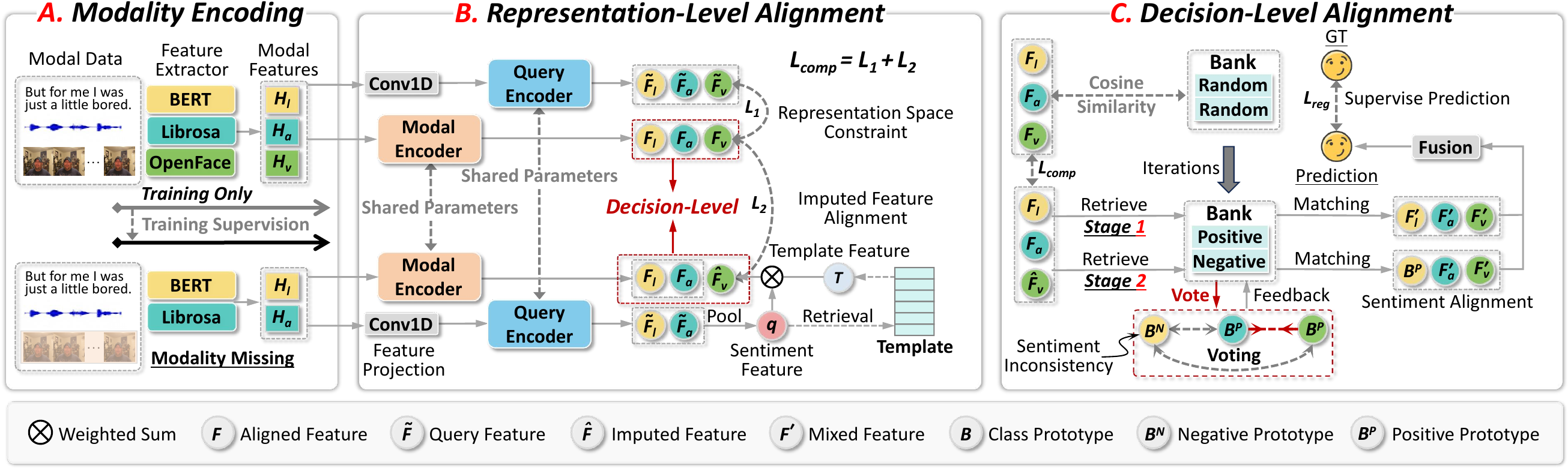}}
		\caption{Overall pipeline of TLRA. The framework consists of three stages: (A) Modality Encoding, (B) Representation-Level Alignment, and (C) Decision-Level Alignment. It first extracts modality-specific features, then aligns incomplete representations via reference-driven completion, and finally performs prototype-guided decision alignment to ensure robust sentiment predictions under modality missing.}
		\label{fig:pipeline}
	\end{figure*}

	\section{Method}
	\subsection{Overview}
	To mitigate both generation bias from modality missing and fusion bias from quality imbalance, we develop the Two-Level Reference Alignment framework (TLRA). TLRA defines cross-modal references at both the representation and decision levels, allowing the model to learn consistent and robust sentiment representations even when only partial or unreliable multimodal inputs are available.
	
	As illustrated in Figure~\ref{fig:pipeline}, TLRA consists of three key components: modality encoding, representation-level alignment, and decision-level alignment. For textual, acoustic, and visual modalities, TLRA independently encodes available modality signals while constructing sentiment reference structures from complete-modality training samples. The representation-level alignment stage constrains incomplete features into a shared sentiment space to reduce generation-induced representation drift. The decision-level alignment stage further enforces cross-modal consistency by suppressing unreliable modality decisions during fusion, thereby reducing fusion bias and improving prediction stability.
	
	\subsection{Modality Encoding}
	In the modality encoding stage, we independently model the raw signals of each modality based on the presence of potentially missing modalities, creating a structured representation for each modality to serve as input for future alignment. Figure~\ref{fig:pipeline}(A) depicts that during the modality encoding stage, the TLRA framework only encodes the available modalities (i.e., does not complete or reconstruct missing modalities) at this stage of the TLRA framework, which allows us to perform the alignment step under true conditions of modality missing.
	
	For the textual, acoustic, and visual modalities, we utilize the BERT pretrained language model, the Librosa acoustic feature extractor, and the OpenFace visual analysis system, respectively, to encode the modalities~\cite{devlin2019bert,mcfee2015librosa,baltruvsaitis2016openface}. A multimodal input will result in three modality representations: $H_l$, $H_a$, and $H_v$ for the text, audio, and visual modalities, respectively.
	
	During training, complete-modality samples are used as supervision references for constructing reference structures. When one or more modalities are missing, TLRA operates only on available modalities without encoding or reconstructing missing inputs. This design allows the model to distinguish missing information from available evidence and avoids introducing unreliable assumptions at the input level.
	
	\subsection{Representation-Level Alignment}
	Imputed features may deviate from the actual sentiment structure and cause incorrect predictions, mainly due to the lack of stable semantic references for anchoring incomplete inputs. TLRA addresses this issue through reference-driven representation-level alignment, which constrains imputed representations within the semantic space defined by complete modalities, as shown in Figure~\ref{fig:pipeline}(B).
	
	\noindent\textbf{Dual-Path Encoding Strategy.} TLRA uses a dual-path encoding strategy to handle both complete-modality and modality-missing samples. The complete-modality path establishes a stable sentiment space from complete inputs, while the modality-missing path serves as a query and completion path for incomplete inputs. By decoupling reference construction from query-based completion, the model can align missing-modality representations within the same sentiment space without assuming that all modalities are available.
	
	Given modality representations $H_m$ from the encoding stage, where $m \in \mathcal{M}$ denotes available modalities, the Modal Encoder maps each modality into a shared sentiment space:
	\begin{equation}
		F_m = \mathrm{Enc}_M(H_m), \quad m \in \mathcal{M}.
	\end{equation}
	
	The resulting features $F_m$ exhibit stable sentiment structure and support subsequent decision-level alignment.
	
	To support representation completion under modality missing, a Query Encoder is introduced to construct query features for reference retrieval. Each available modality is first projected into a unified representation space via a lightweight 1D convolution, and then processed by the Query Encoder:
	\begin{equation}
		\tilde F_m = \mathrm{Enc}_Q\bigl(\mathrm{Conv1D}(H_m)\bigr), \quad m \in \mathcal{M}.
	\end{equation}
	
	All modalities pass through the query path regardless of completeness, producing features that support both completion and alignment. By decoupling reference construction from query-based alignment in a shared semantic space, the dual-path design enables robust alignment without assuming complete inputs. The two-path encoding is implemented with shared lightweight modules to ensure efficiency and stability.
	
	\noindent\textbf{Template-Guided Completion.}
	Under modality missing, unconstrained feature generation can produce biased representations deviating from true sentiment structures. To address this, TLRA introduces a template-guided completion mechanism that provides stable semantic references for missing modalities and constrains completion within the sentiment space defined by complete modalities.
	
	Given query features $\tilde F_m$ from the Query Encoding, we apply temporal pooling to obtain a global query vector $q_m$:
	\begin{equation}
		q_m = \mathrm{Pool}(\tilde F_m).
	\end{equation}
	
	TLRA maintains a learnable template library $\mathcal{T}=\{t_1, t_2, \dots, t_K\}$ in the shared sentiment space. The similarity between $q_m$ and each template is computed as:
	\begin{equation}
		s_k = {q_m^{\rm{T}}} t_k.
	\end{equation}
	
	The similarities are normalized via softmax to obtain aggregation weights $\alpha_k$, yielding the template context as:
	\begin{equation}
		T_m = \sum_{k=1}^{K} \alpha_k t_k.
	\end{equation}
	
	To adaptively fuse the sample-specific query and the template reference, we compute a consistency weight based on their cosine similarity as:
	\begin{equation}
		\alpha_c = \cos(q_m, T_m),
	\end{equation}
	
	The completed representation is then obtained as:
	\begin{equation}
		\hat F_m = \alpha_c\, q_m + (1-\alpha_c)\, T_m.
	\end{equation}
	
	The completed representation $\hat F_m$ serves as sentiment-level structural compensation rather than reconstruction of the original modality signal. Therefore, the completion process focuses on recovering stable sentiment structure instead of synthesizing raw modality-specific details, which reduces the risk of introducing modality-specific generation noise. By constraining completed features within the shared sentiment space, template-guided completion effectively suppresses generation bias under modality missing and provides stable inputs for alignment and fusion.
	
	\noindent\textbf{Representation-Level Alignment Loss.}
	The objective of representation-level alignment is to enforce structural consistency among intermediate representations within a unified sentiment space under modality missing. Specifically, both query features for observed modalities and completed features for missing modalities are aligned to the modality representations defined under complete inputs.
	
	Let $\mathcal{M}$ denote the set of available modalities and $\mathcal{M}_{\text{miss}}$ the set of missing modalities. We define $\Omega = \mathcal{M} \cup \mathcal{M}_{\text{miss}}$ as the set of modalities involved in alignment. Let $F_m$ denote the modality feature produced by the modal encoder, $\tilde F_m$ the query feature, and $\hat F_m$ the completed feature. The unified representation alignment loss is defined as:
	\begin{equation}
		\mathcal{L}_{\text{align}}
		=
		\lambda_1 \sum_{m \in \mathcal{M}}
		\left\|
		\tilde F_m - F_m
		\right\|_2^2
		+
		\lambda_2 \sum_{m \in \mathcal{M}_{\text{miss}}}
		\left\|
		\hat F_m - F_m
		\right\|_2^2 .
	\end{equation}

	By aligning query features and completed missing-modality features to complete-modality references, this loss suppresses representation drift caused by missing inputs. As a result, different modality combinations are encouraged to follow a consistent sentiment structure, providing more reliable inputs for decision-level alignment.
	
	\subsection{Decision-Level Alignment}
	Although representation-level alignment reduces missing-induced drift, decision inconsistencies may still occur when noisy modalities produce misleading predictions. These unreliable modality-wise predictions can dominate fusion and degrade the final sentiment decision. As shown in Figure~\ref{fig:pipeline}(C), TLRA filters and aligns modality-wise predictions to suppress unreliable evidence under varying modality conditions and improve decision robustness.
	
	\noindent\textbf{Prototype Construction.}
	For each sentiment class $c$, we maintain modality-specific prototypes $B_m^{c}$ located within the shared sentiment space for the text, audio, and visual modalities. These prototypes serve as decision-level references and are constructed based upon accumulated training features, capturing stable class structural elements across modalities. In practice, the prototype bank is randomly initialized and then updated online during training, so no extra clustering procedure or external prior is required.
	
	Given the aligned feature $F_m$ of the sample with class label $c$, we determine the cosine similarity with the associated prototype $B_m^{c}$, which is utilized to create a mixed prototype:
	
	\begin{equation}
		{B'}_m^{c} = \alpha_m F_m + (1-\alpha_m) B_m^{c},
	\end{equation}
	where $\alpha_m$ is determined by the cosine similarity between $F_m$ and $B_m^{c}$, adaptively balancing sample-specific information and class-level structure to improve robustness.

\begin{figure}[!t]
		\centering{\includegraphics[width=1\linewidth]{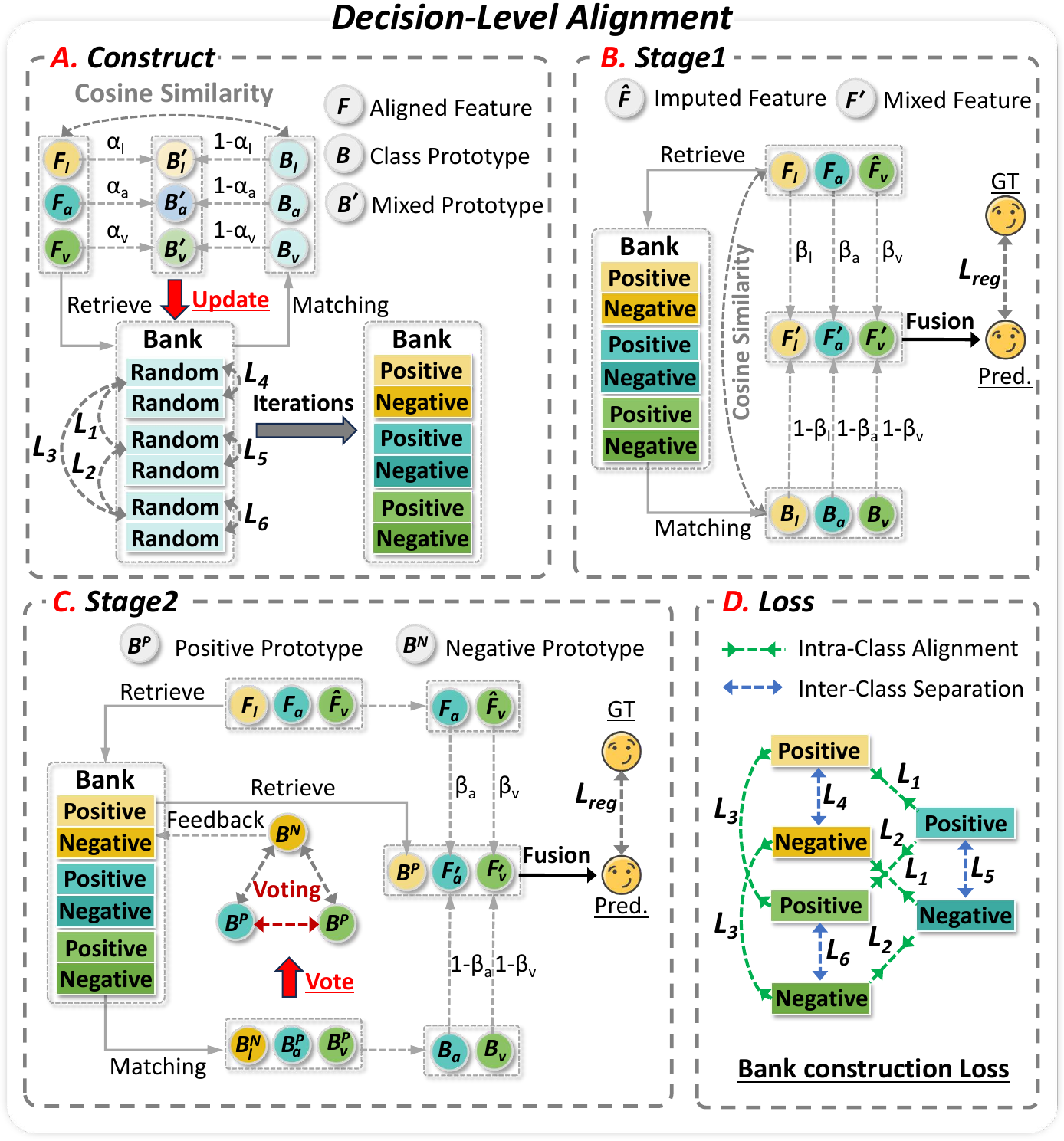}}
		\caption{Decision-level alignment in TLRA. It illustrates prototype construction, a two-stage decision alignment strategy, and the training objectives for consistent and robust sentiment predictions.}
		\label{fig:memo}
	\end{figure}

	The mixed prototype ${B'}_m^{c}$ enables gradual updates of the prototype memory with controlled momentum, allowing prototypes to evolve into stable class-level references while remaining robust to sample variations, as illustrated in Figure~\ref{fig:memo}(A). In the binary sentiment setting, each modality maintains one positive and one negative prototype, keeping the memory compact and interpretable while providing explicit class-level anchors. This class-wise design avoids additional prototype-number search and ensures that decision references are directly aligned with sentiment labels. During training, we adopt momentum-based updates with normalization to stabilize cosine similarity, enabling prototypes to progressively evolve from coarse initial references into reliable and discriminative sentiment anchors in practice.
	
	Through this process, class prototypes act as decision-level anchors that provide consistent references across samples and modalities. As the prototype memory is continuously updated during training, it gradually captures stable class-level sentiment structures and helps resolve cross-modal prediction conflicts under noisy and unreliable modality conditions.
	
	\noindent\textbf{Two-Stage Decision-Level Alignment.}
	Since prototype memory gradually evolves from coarse class representations into reliable references, TLRA adopts a two-stage decision-level alignment strategy that progressively strengthens cross-modal consistency, as shown in Figure~\ref{fig:pipeline}(C).
	
	\textit{Stage1: Soft Prototype Guidance.}
	At early epochs, prototype memory is not yet stable enough to support hard voting, and strong cross-modal constraints may amplify modality noise. Therefore, prototypes provide continuous sentiment references without explicit voting or hard class decisions. This soft guidance preserves useful modality-specific information while gradually pulling unstable features toward smoother and more reliable prototype structures.

	Given the aligned modality feature $F_m$ for each available modality $m \in \mathcal{M}$, the model retrieves the corresponding positive and negative prototypes $B_m^{P}$ and $B_m^{N}$ from the prototype memory and computes their cosine similarities:
	\begin{equation}
		s_m^{P}=\cos(F_m,B_m^{P}),\qquad
		s_m^{N}=\cos(F_m,B_m^{N}).
	\end{equation}
	
	The prototype with higher similarity is selected as the primary sentiment reference for the current modality:
	\begin{equation}
		B_m^{*}=\argmax_{c\in\{P,N\}} \cos(F_m,B_m^{c}),
	\end{equation}
	and the maximum similarity is used as a consistency score:
	\begin{equation}
		\beta_m=\max(s_m^{P},\,s_m^{N}).
	\end{equation}
	
	Based on this score, the modality feature is fused with the selected prototype to form a mixed representation $F_m'$:
	\begin{equation}
		F_m'=\beta_m F_m+(1-\beta_m)B_m^{*}.
	\end{equation}
	
	This soft guidance preserves reliable modality information while pulling unstable modalities toward smoother prototype structures. The resulting modality representations are then fused to produce the final sentiment prediction.
	
	\textit{Stage2: Prototype Voting and Suppression.}
	As training progresses, the prototype memory gradually becomes compact and discriminative. In the second stage, TLRA introduces an explicit prototype voting mechanism to enforce cross-modal decision consistency, as shown in Figure~\ref{fig:pipeline}(C). Each modality produces a prototype-space decision as:
	\begin{equation}
		v_m=\argmax_{c\in\{P,N\}} s_m^{c}.
	\end{equation}
	
	The modality-wise decisions $v_m$ form a voting process, where the majority vote defines the high-confidence sentiment direction. Disagreeing modalities are suppressed and replaced with prototype features consistent with the voted sentiment, preventing unreliable modalities from dominating fusion. The voting mechanism is activated only in the second stage, allowing the first stage to stabilize prototype formation before stronger decision-level consistency is enforced.
	
	\noindent\textbf{Decision-Level Alignment Loss.}
	Decision-level alignment aims to construct a discriminative and cross-modally consistent sentiment prototype space, and to constrain multimodal fusion decisions to align with ground-truth sentiment labels. To this end, TLRA introduces three loss terms at the decision level: two for prototype structure learning and one for final sentiment prediction to further enhance robustness under modality missing and quality imbalance.
	
	First, to ensure intrinsic discriminability within each modality, the positive (P) and negative (N) sentiment prototypes should be well separated in the prototype space. Let $B_m^P$ and $B_m^N$ denote the positive and negative prototypes of modality $m$. We define an intra-modality separation loss as:
	
	\begin{equation}
		\mathcal{L}_{\text{intra}}
		=
		\sum_{m \in \mathcal{M}}
		\max\!\bigl(0,\; \delta - \cos(B_m^P, B_m^N)\bigr),
	\end{equation}
	where $\delta$ is a margin threshold enforcing clear class boundaries within each modality to enhance separability.
	
	Second, to promote cross-modal consistency, prototypes corresponding to the same sentiment across different modalities are encouraged to align in the sentiment space. The cross-modal prototype alignment loss is defined as:
	\begin{equation}
		\mathcal{L}_{\text{inter}}
		=
		\sum_{\substack{m,n \in \mathcal{M}\\ m \neq n}}
		\Bigl(
		1 - \cos(B_m^P, B_n^P)
		+
		1 - \cos(B_m^N, B_n^N)
		\Bigr).
	\end{equation}
	
	\begin{table*}[!t]
		\centering
			\includegraphics[width=\textwidth]{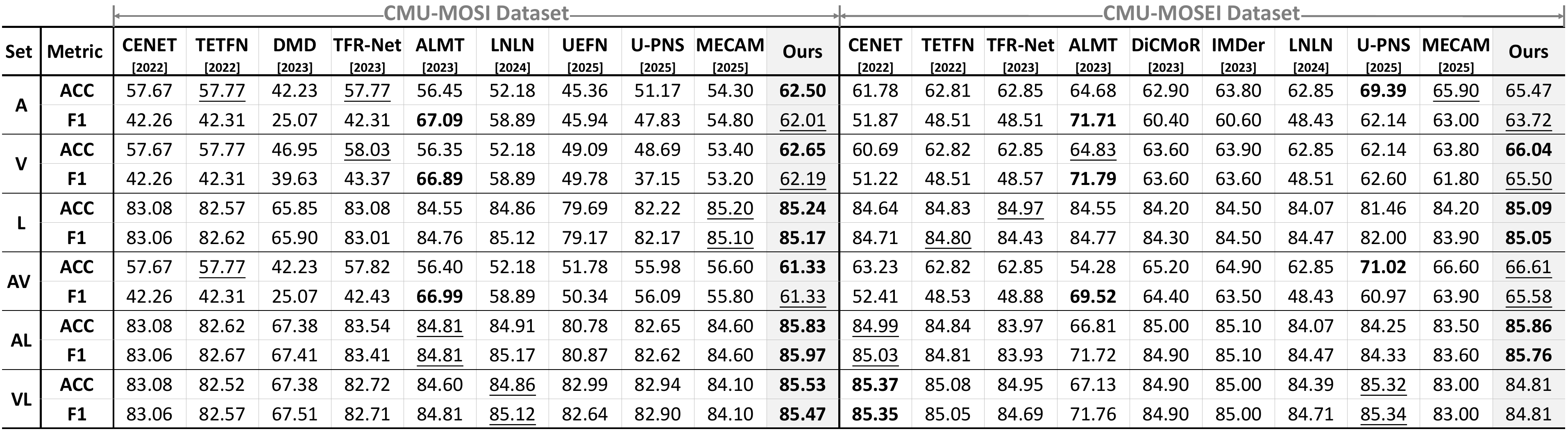}
		\caption{Quantitative comparison with representative state-of-the-art methods under missing-modality settings on the CMU-MOSI and CMU-MOSEI datasets. The best results in each column are highlighted in \textbf{bold}, and the second-best results are underlined. A, V, and L denote the audio, visual, and language modalities, respectively.}
		\label{tab:miss}
	\end{table*}
	
	After prototype construction and alignment, the final sentiment prediction is obtained through the fusion module. A standard task supervision loss is applied:
	\begin{equation}
		\mathcal{L}_{\text{reg}} = \ell(\hat{y}, y),
	\end{equation}
	where $\hat{y}$ and $y$ denote the predicted and ground-truth sentiment labels, respectively, for each input sample. The overall decision-level optimization objective is:
	\begin{equation}
		\mathcal{L}_{\text{total}}
		=
		\lambda_3 \mathcal{L}_{\text{intra}}
		+
		\lambda_4 \mathcal{L}_{\text{inter}}
		+
		\lambda_5 \mathcal{L}_{\text{reg}}
		+
		\mathcal{L}_{\text{align}} .
	\end{equation}
	
	Together, these losses enforce intra-modality discriminability, cross-modal consistency, and prediction correctness, enabling TLRA to produce robust sentiment predictions under modality quality imbalance. The loss weighting coefficients are selected via controlled validation, and the model remains stable across a reasonable range of settings, indicating that the performance gain primarily arises from the proposed representation- and decision-level alignment mechanisms rather than sensitive hyperparameter tuning.
	
	\begin{table}[!t]
		\centering
			\includegraphics[width=1\linewidth]{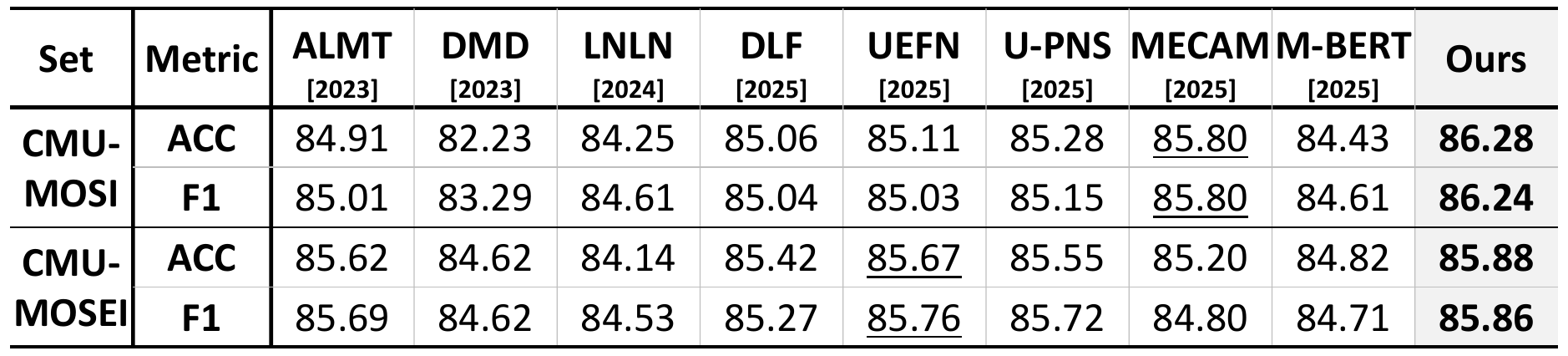}
		\caption{Quantitative comparison with representative SOTA methods under full-modality settings on the CMU-MOSI and CMU-MOSEI datasets. The best results in each column are highlighted in \textbf{bold}, and the second-best results are underlined.}
		\label{tab:full}
	\end{table}
	
	\section{Experiments}
	\subsection{Experimental Setup, Datasets, and Metrics}
	\noindent\textbf{Experimental Setup.}
	We trained our model using PyTorch and 1 NVIDIA RTX 4090 GPU.
	Each modality encoder is initialized with a pretrained model, and alignment and fusion modules were learned end-to-end. Using AdamW with a learning rate of $1 \times 10^{-4}$ and a cosine annealing schedule, the batch size is 16 and was trained over 100 epochs. All hyperparameters were kept constant across all experiments, where $\lambda_1=\lambda_2=1.0$, $\lambda_3=\lambda_4=0.7$, and $\lambda_5=0.2$. These coefficients are selected by one-factor-at-a-time validation to favor settings that remain stable across modality combinations rather than relying on exhaustive search. The additional cost of TLRA mainly comes from template retrieval and prototype matching, which scale linearly with the number of templates and prototypes. Since TLRA does not perform input-level modality reconstruction, the dominant computation remains in modality encoding, and the alignment modules introduce only limited overhead. This indicates that TLRA improves robustness without relying on expensive generation or reconstruction modules, making it more practical for deployment under resource-constrained settings.
	
	\noindent\textbf{Datasets.}
	We assessed our methods against two publicly available benchmarks for multimodal sentiment analysis, CMU-MOSI~\cite{zadeh2016mosi} and CMU-MOSEI~\cite{zadeh2018multimodal}. For both datasets the videos and modal data were aligned; CMU-MOSI consists of 2199 utterances from 93 video clips, while CMU-MOSEI is a larger dataset containing 23,453 utterances from over 1000 speakers. With aligned multimodal annotations, we were able to perform end-to-end fusion learning with these datasets, as well as evaluate our algorithms under missing-modality settings.
	
	\noindent\textbf{Metrics.}
	The performance metrics for both datasets were assessed using ACC and F1-score (binary sentiment), where samples were assigned to the negative class or the non-negative class based on standard thresholding of sentiment scores. ACC measures overall classification accuracy and F1-score complements ACC by balancing precision and recall.
	
	\subsection{Quantitative Evaluations}
	\noindent\textbf{Missing-Modality Comparison.}
	Table~\ref{tab:miss} shows a detailed quantitative comparison between our approach under missing-modality scenarios on both CMU-MOSI and CMU-MOSEI datasets, in addition to comparing other representative approaches for multimodal sentiment analysis under similar conditions of missing modalities, such as CENET~\cite{wang2022cross}, TETFN~\cite{wang2023tetfn}, DMD~\cite{li2023decoupled}, TFR-Net~\cite{yuan2021transformer}, ALMT~\cite{zhang2023learning}, LNLN~\cite{zhang2024towards}, UEFN~\cite{wang2025uefn}, U-PNS~\cite{chen2025ucmib}, MECAM~\cite{wang2025contrastive}, DiCMoR~\cite{wang2023distribution}, and IMDer~\cite{wang2023incomplete}. All methods are evaluated using ACC and F1-score across all conditions for binary sentiment classification.
	
	Our approach performs better or remains competitive across most modality combinations, showing that TLRA mitigates degradation from incomplete multimodal information. The gains are clearer under severe missing settings, indicating stronger robustness with limited evidence. Representation-level alignment stabilizes incomplete features, while decision-level alignment reduces unreliable modality-wise predictions. Together, they address representation drift and decision inconsistency, producing more reliable sentiment predictions. Since missing-modality and reliability-imbalance settings both reflect modality reliability degradation, TLRA can handle unreliable evidence without relying on specific noise patterns, highlighting its adaptability to real-world conditions and varying data quality.
	
	\noindent\textbf{Full-Modality Comparison.}
	Quantitative performance comparisons for our full-modality evaluation condition can be found in Table~\ref{tab:full} where we compare our method to other representative multimodal sentiment analysis methods including ALMT, DMD, LNLN, DLF~\cite{wang2025dlf}, UEFN, U-PNS and M-BERT~\cite{rahman2020integrating}.
	
	Overall, the proposed framework achieves the highest ACC and F1-score on both CMU-MOSI and CMU-MOSEI, demonstrating robustness under complete-modality settings. Beyond performance gains, this suggests TLRA improves multimodal fusion even when all modalities are present. This also indicates TLRA is not limited to missing-modality scenarios; its decision-level reference mechanism improves fusion reliability when modalities exhibit inconsistent quality or conflicting sentiment cues. This further highlights the effectiveness of reference-guided decision alignment in promoting consistent and reliable multimodal predictions.
	
	\begin{table}[!t]
		\centering
			\includegraphics[width=1\linewidth]{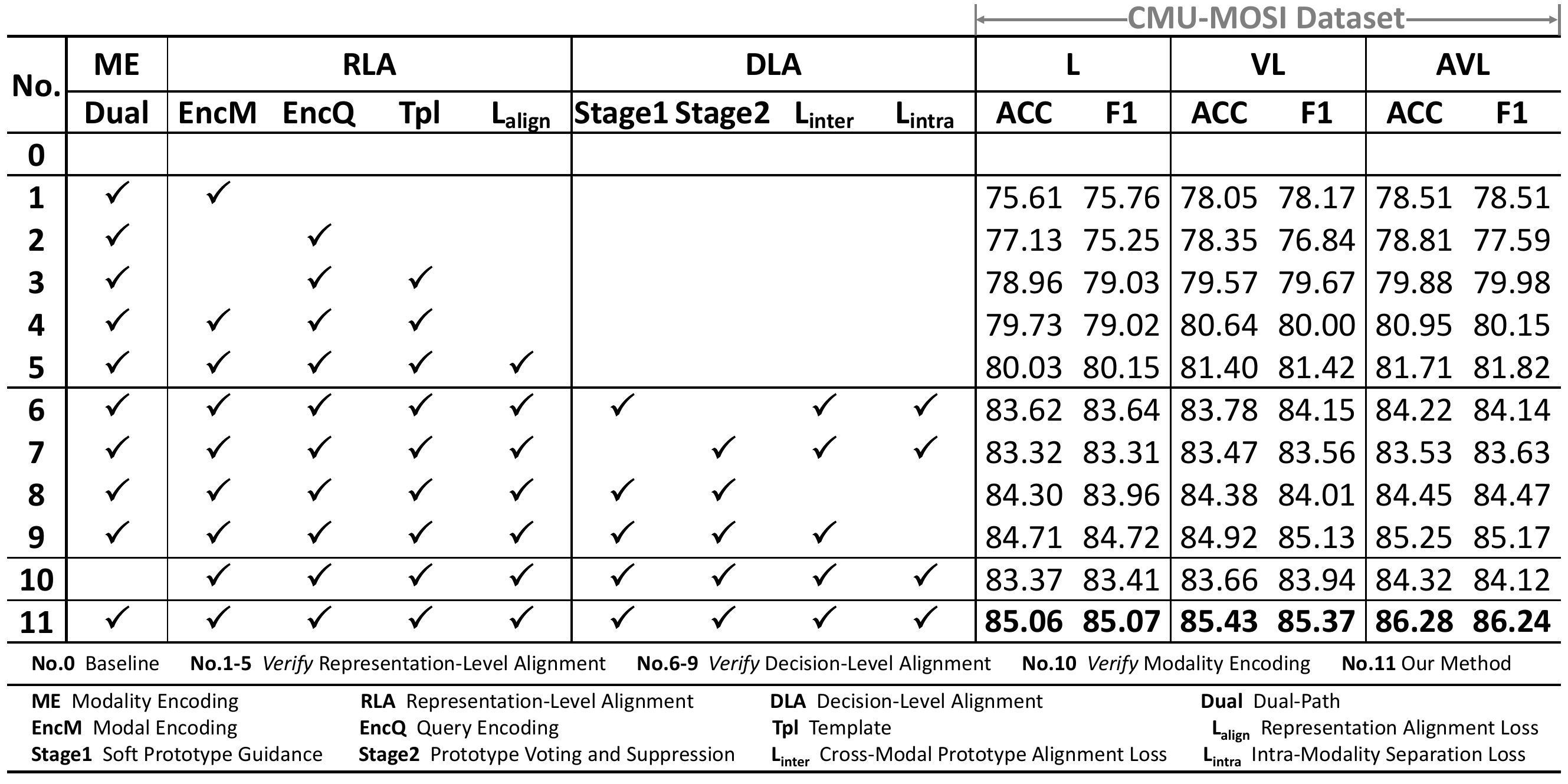}
		\caption{Ablation study of Modality Encoding, Representation-Level Alignment, and Decision-Level Alignment components on the CMU-MOSI dataset under different modality settings. }
		\label{tab:comp}
	\end{table}
	
	\subsection{Ablation Experiments}
	We conduct systematic ablation studies on CMU-MOSI, with results summarized in Table~\ref{tab:comp}. By progressively adding modality encoding (ME), representation-level alignment (RLA), and decision-level alignment (DLA) to a baseline, we isolate the contribution of each component.
	
	Results from No.~1 to No.~5 show that RLA consistently improves performance, indicating that stable references effectively mitigate representation drift under modality missing. Results from No.~6 to No.~9 further confirm the effectiveness of DLA; notably, the two-stage strategy yields consistent gains by suppressing unreliable modalities via prototype guidance and voting to enhance cross-modal consistency.
	
	The full model (No.~11) achieves the best performance across all modality settings, demonstrating that jointly aligning representations and decisions is critical for robustness under modality-missing conditions.

	\subsection{Cross-Dataset Generalization}
	Table~\ref{tab:cross} reports the cross-dataset evaluation results, where models are trained on one dataset and tested on the other. This setting is particularly challenging because speakers, topics, and expression styles vary significantly across datasets, often leading to overfitting on dataset-specific patterns. Compared with representative baselines, our method consistently achieves the best performance and exhibits smaller degradation when transferring across datasets. This indicates that TLRA effectively reduces reliance on dataset-specific cues and captures more intrinsic and transferable sentiment structures. As a result, TLRA learns more stable and dataset-invariant sentiment representations, enabling robust generalization under distribution shifts and unseen data conditions.
	\begin{table}[!t]
		\centering
			\includegraphics[width=1\linewidth]{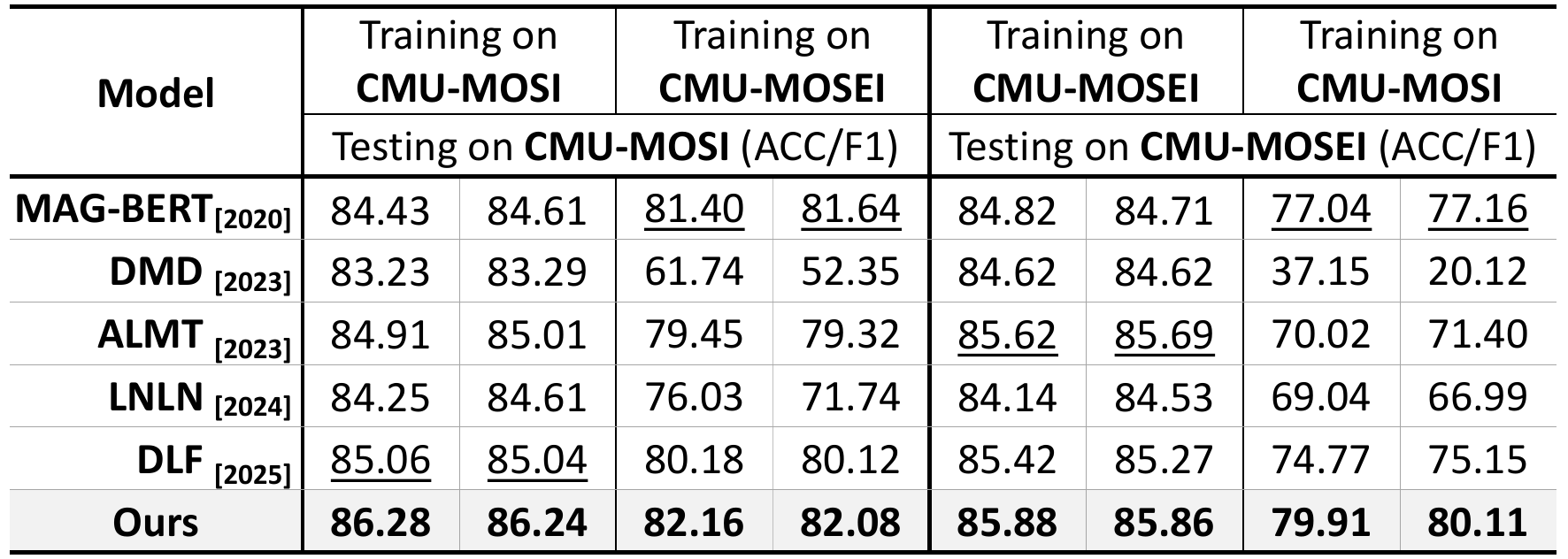}
		\caption{Cross-dataset evaluation results on CMU-MOSI and CMU-MOSEI, where models are trained on one dataset and tested on the other. This setting evaluates generalization under mismatch.}
		\label{tab:cross}
	\end{table}
	
	\begin{figure}[!t]
		\centering{\includegraphics[width=1\linewidth]{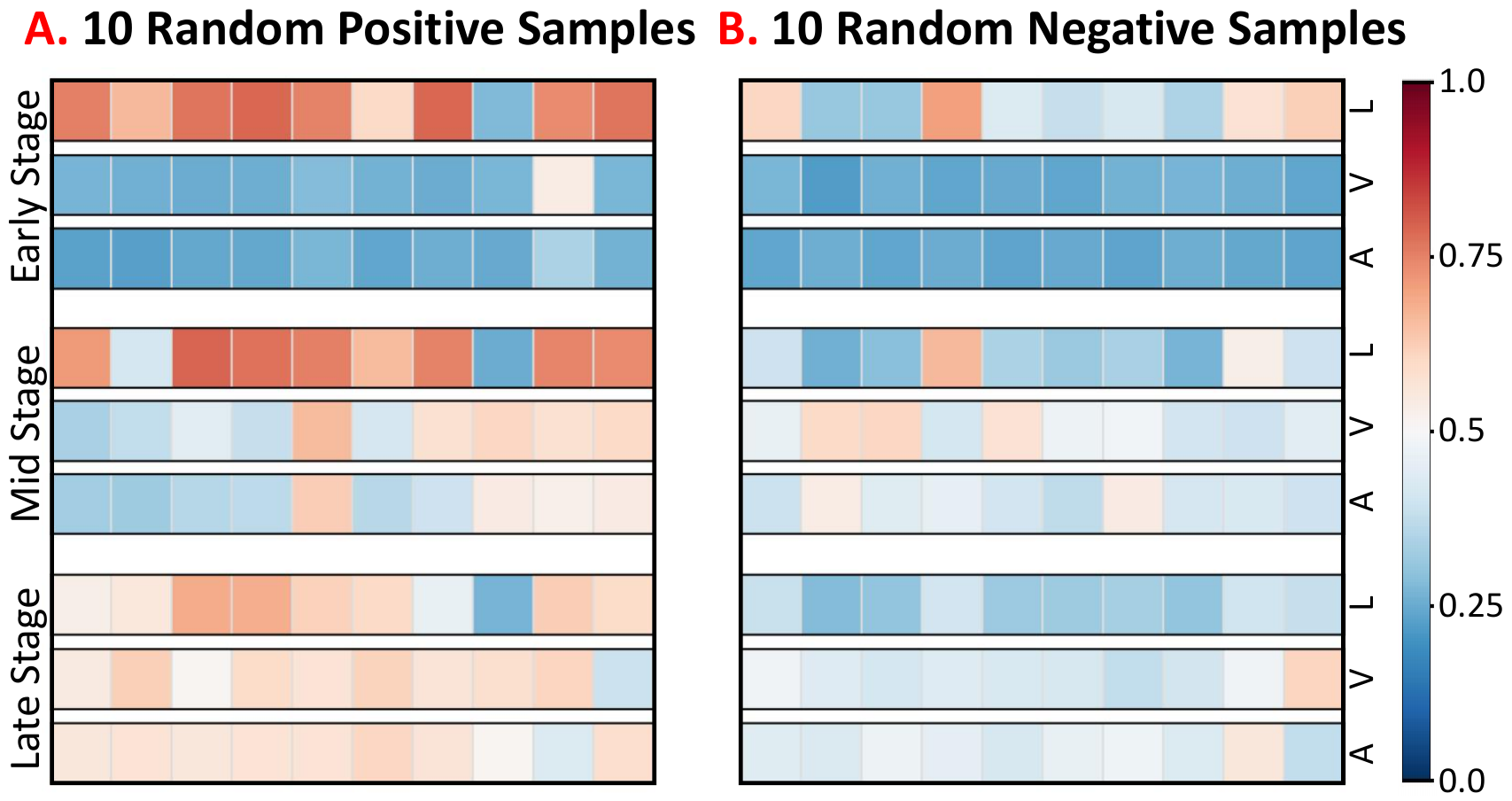}}
		\caption{Similarity difference visualization between randomly sampled sentiment samples and sentiment prototypes across training stages. Lighter colors indicate improved cross-modal alignment and more consistent sentiment representations.}
		\label{fig:sim}
	\end{figure}
	
	\subsection{Prototype Similarity Analysis}
	Figure~\ref{fig:sim} visualizes similarity differences between selected samples and their prototypes across training stages. We select ten positive and ten negative samples and evaluate them at 10, 40, and 70 epochs. In early stages, high contrast indicates modality-specific differences, suggesting that the feature space is dominated by modality variations and inconsistent sentiment representations. As training proceeds, the colors become lighter and more uniform, showing that samples align with their prototypes and converge toward a shared sentiment structure. At later stages, samples show consistent prototype similarity, indicating that the prototype bank has become stable class-level anchors. This trend supports the two-stage strategy, where soft guidance stabilizes the prototype space before voting-based suppression is applied.
	
	\section{Conclusion}	
	We present the Two-Level Reference Alignment framework to address decision drift in multimodal sentiment analysis under missing or unreliable modality conditions. TLRA aligns feature representations and final decisions to stable references, mitigating representation deviation and biased fusion caused by incomplete or unreliable evidence. Experiments on CMU-MOSI and CMU-MOSEI show robust performance across missing-modality, full-modality, and cross-dataset settings. These results demonstrate that feature- and decision-level reference constraints improve stability, robustness, and generalization under diverse modality degradation scenarios, while reducing sensitivity to dataset-specific sentiment patterns and modality inconsistencies. The framework is general and can be extended to other multimodal learning tasks.
	
	\section*{Acknowledgments}
	This work was supported by the National Natural Science Foundation of China (No. 62172246), the Excellent Young Scientists Fund of the Natural Science Foundation of Shandong Province (No. ZR2024YQ071), the Key Laboratory of Forensic Examination for Sichuan Provincial Universities (No. 2024YB01), and the Fundamental Research Funds for the Central Universities (No. 22CX06037A).
	
	\bibliographystyle{named}
	\bibliography{ijcai26}

@inproceedings{yu2021learning,
	title={Learning Modality-Specific Representations with Self-Supervised Multi-Task Learning for Multimodal Sentiment Analysis},
	author={Yu, Wenmeng and Xu, Hua and Yuan, Ziqi and Wu, Jiele},
	booktitle={AAAI},
	volume={35},
	number={12},
	pages={10790--10797},
	year={2021}
}

@book{liu2022sentiment,
	title={Sentiment analysis and opinion mining},
	author={Liu, Bing},
	year={2022},
	publisher={Springer Nature}
}

@article{baltruvsaitis2018multimodal,
	title={Multimodal machine learning: A survey and taxonomy},
	author={Baltru{\v{s}}aitis, Tadas and Ahuja, Chaitanya and Morency, Louis-Philippe},
	journal={IEEE TPAMI},
	volume={41},
	number={2},
	pages={423--443},
	year={2018},
	publisher={IEEE}
}

@article{zadeh2016mosi,
	title={Mosi: multimodal corpus of sentiment intensity and subjectivity analysis in online opinion videos},
	author={Zadeh, Amir and Zellers, Rowan and Pincus, Eli and Morency, Louis-Philippe},
	journal={arXiv preprint arXiv:1606.06259},
	year={2016}
}

@inproceedings{zadeh2018multimodal,
	title={Multimodal language analysis in the wild: Cmu-mosei dataset and interpretable dynamic fusion graph},
	author={Zadeh, AmirAli Bagher and Liang, Paul Pu and Poria, Soujanya and Cambria, Erik and Morency, Louis-Philippe},
	booktitle={Proceedings of the 56th Annual Meeting of the Association for Computational Linguistics (Volume 1: Long Papers)},
	pages={2236--2246},
	year={2018}
}

@article{neverova2015moddrop,
	title={Moddrop: adaptive multi-modal gesture recognition},
	author={Neverova, Natalia and Wolf, Christian and Taylor, Graham and Nebout, Florian},
	journal={IEEE TPAMI},
	volume={38},
	number={8},
	pages={1692--1706},
	year={2015},
	publisher={IEEE}
}

@article{arevalo2017gated,
	title={Gated multimodal units for information fusion},
	author={Arevalo, John and Solorio, Thamar and Montes-y-G{\'o}mez, Manuel and Gonz{\'a}lez, Fabio A},
	journal={arXiv preprint arXiv:1702.01992},
	year={2017}
}

@inproceedings{pham2019found,
	title={Found in translation: Learning robust joint representations by cyclic translations between modalities},
	author={Pham, Hai and Liang, Paul Pu and Manzini, Thomas and Morency, Louis-Philippe and P{\'o}czos, Barnab{\'a}s},
	booktitle={AAAI},
	volume={33},
	number={01},
	pages={6892--6899},
	year={2019}
}

@article{lin2023missmodal,
	title={Missmodal: Increasing robustness to missing modality in multimodal sentiment analysis},
	author={Lin, Ronghao and Hu, Haifeng},
	journal={TACL},
	volume={11},
	pages={1686--1702},
	year={2023},
	publisher={MIT Press One Broadway, 12th Floor, Cambridge, Massachusetts 02142, USA~…}
}

@inproceedings{liu2018efficient,
	title={Efficient low-rank multimodal fusion with modality-specific factors},
	author={Liu, Zhun and Shen, Ying and Lakshminarasimhan, Varun Bharadhwaj and Liang, Paul Pu and Zadeh, AmirAli Bagher and Morency, Louis-Philippe},
	booktitle={ACL},
	pages={2247--2256},
	year={2018}
}

@inproceedings{tsai2019multimodal,
	title={Multimodal transformer for unaligned multimodal language sequences},
	author={Tsai, Yao-Hung Hubert and Bai, Shaojie and Liang, Paul Pu and Kolter, J Zico and Morency, Louis-Philippe and Salakhutdinov, Ruslan},
	booktitle={ACL},
	volume={2019},
	pages={6558},
	year={2019}
}

@article{han2021improving,
	title={Improving multimodal fusion with hierarchical mutual information maximization for multimodal sentiment analysis},
	author={Han, Wei and Chen, Hui and Poria, Soujanya},
	journal={arXiv preprint arXiv:2109.00412},
	year={2021}
}

@inproceedings{hazarika2020misa,
	title={Misa: Modality-invariant and-specific representations for multimodal sentiment analysis},
	author={Hazarika, Devamanyu and Zimmermann, Roger and Poria, Soujanya},
	booktitle={ACM MM},
	pages={1122--1131},
	year={2020}
}

@inproceedings{zadeh2018memory,
	title={Memory fusion network for multi-view sequential learning},
	author={Zadeh, Amir and Liang, Paul Pu and Mazumder, Navonil and Poria, Soujanya and Cambria, Erik and Morency, Louis-Philippe},
	booktitle={AAAI},
	volume={32},
	number={1},
	year={2018}
}

@inproceedings{wang2019words,
	title={Words can shift: Dynamically adjusting word representations using nonverbal behaviors},
	author={Wang, Yansen and Shen, Ying and Liu, Zhun and Liang, Paul Pu and Zadeh, Amir and Morency, Louis-Philippe},
	booktitle={AAAI},
	volume={33},
	number={01},
	pages={7216--7223},
	year={2019}
}

@inproceedings{devlin2019bert,
	title={Bert: Pre-training of deep bidirectional transformers for language understanding},
	author={Devlin, Jacob and Chang, Ming-Wei and Lee, Kenton and Toutanova, Kristina},
	booktitle={NAACL-HLT},
	pages={4171--4186},
	year={2019}
}

@article{mcfee2015librosa,
	title={librosa: Audio and music signal analysis in python.},
	author={McFee, Brian and Raffel, Colin and Liang, Dawen and Ellis, Daniel PW and McVicar, Matt and Battenberg, Eric and Nieto, Oriol},
	journal={SciPy},
	volume={2015},
	pages={18--24},
	year={2015}
}

@inproceedings{baltruvsaitis2016openface,
	title={Openface: an open source facial behavior analysis toolkit},
	author={Baltru{\v{s}}aitis, Tadas and Robinson, Peter and Morency, Louis-Philippe},
	booktitle={WACV},
	pages={1--10},
	year={2016},
	organization={IEEE}
}

@article{wang2022cross,
	title={Cross-modal enhancement network for multimodal sentiment analysis},
	author={Wang, Di and Liu, Shuai and Wang, Quan and Tian, Yumin and He, Lihuo and Gao, Xinbo},
	journal={TMM},
	volume={25},
	pages={4909--4921},
	year={2022},
	publisher={IEEE}
}

@article{wang2023tetfn,
	title={TETFN: A text enhanced transformer fusion network for multimodal sentiment analysis},
	author={Wang, Di and Guo, Xutong and Tian, Yumin and Liu, Jinhui and He, LiHuo and Luo, Xuemei},
	journal={PR},
	volume={136},
	pages={109259},
	year={2023},
	publisher={Elsevier}
}

@inproceedings{li2023decoupled,
	title={Decoupled multimodal distilling for emotion recognition},
	author={Li, Yong and Wang, Yuanzhi and Cui, Zhen},
	booktitle={CVPR},
	pages={6631--6640},
	year={2023}
}

@inproceedings{yuan2021transformer,
	title={Transformer-based feature reconstruction network for robust multimodal sentiment analysis},
	author={Yuan, Ziqi and Li, Wei and Xu, Hua and Yu, Wenmeng},
	booktitle={ACM MM},
	pages={4400--4407},
	year={2021}
}

@article{zhang2023learning,
	title={Learning language-guided adaptive hyper-modality representation for multimodal sentiment analysis},
	author={Zhang, Haoyu and Wang, Yu and Yin, Guanghao and Liu, Kejun and Liu, Yuanyuan and Yu, Tianshu},
	journal={arXiv preprint arXiv:2310.05804},
	year={2023}
}

@article{zhang2024towards,
	title={Towards robust multimodal sentiment analysis with incomplete data},
	author={Zhang, Haoyu and Wang, Wenbin and Yu, Tianshu},
	journal={NIPS},
	volume={37},
	pages={55943--55974},
	year={2024}
}

@article{wang2025uefn,
	title={UEFN: Efficient uncertainty estimation fusion network for reliable multimodal sentiment analysis},
	author={Wang, Shuai and Ratnavelu, K and Bin Shibghatullah, Abdul Samad},
	journal={Applied Intelligence},
	volume={55},
	number={3},
	pages={171},
	year={2025},
	publisher={Springer}
}

@article{chen2025ucmib,
	title={UCMIB-PNS: Balancing Sufficiency and Necessity with Probabilistic Causality and Cross-Modal Uncertainty in Multimodal Sentiment Analysis},
	author={Chen, Jili and Zhong, Yihua and Huang, Qionghao and Huang, Changqin and Jiang, Fan and Huang, Xiaodi and Wang, Xun},
	journal={IEEE TAC},
	year={2025},
	publisher={IEEE}
}

@article{wang2025contrastive,
	title={Contrastive-Based Removal of Negative Information in Multimodal Emotion Analysis},
	author={Wang, Rui and Wang, Yaoyang and Cambria, Erik and Fan, Xuhui and Yu, Xiaohan and Huang, Yao and E, Xiaosong and Zhu, Xianxun},
	journal={Cognitive Computation},
	volume={17},
	number={3},
	pages={107},
	year={2025},
	publisher={Springer}
}

@inproceedings{wang2023distribution,
	title={Distribution-consistent modal recovering for incomplete multimodal learning},
	author={Wang, Yuanzhi and Cui, Zhen and Li, Yong},
	booktitle={ICCV},
	pages={22025--22034},
	year={2023}
}

@article{wang2023incomplete,
	title={Incomplete multimodality-diffused emotion recognition},
	author={Wang, Yuanzhi and Li, Yong and Cui, Zhen},
	journal={NIPS},
	volume={36},
	pages={17117--17128},
	year={2023}
}

@inproceedings{rahman2020integrating,
	title={Integrating multimodal information in large pretrained transformers},
	author={Rahman, Wasifur and Hasan, Md Kamrul and Lee, Sangwu and Zadeh, AmirAli Bagher and Mao, Chengfeng and Morency, Louis-Philippe and Hoque, Ehsan},
	booktitle={ACL},
	pages={2359--2369},
	year={2020}
}

@inproceedings{wang2025dlf,
	title={DLF: Disentangled-language-focused multimodal sentiment analysis},
	author={Wang, Pan and Zhou, Qiang and Wu, Yawen and Chen, Tianlong and Hu, Jingtong},
	booktitle={AAAI},
	volume={39},
	number={20},
	pages={21180--21188},
	year={2025}
}

@article{poria2017review,
	title={A review of affective computing: From unimodal analysis to multimodal fusion},
	author={Poria, Soujanya and Cambria, Erik and Bajpai, Rajiv and Hussain, Amir},
	journal={Information fusion},
	volume={37},
	pages={98--125},
	year={2017},
	publisher={Elsevier}
}

@inproceedings{zadeh2018multi,
	title={Multi-attention recurrent network for human communication comprehension},
	author={Zadeh, Amir and Liang, Paul Pu and Poria, Soujanya and Vij, Prateek and Cambria, Erik and Morency, Louis-Philippe},
	booktitle={AAAI},
	volume={32},
	number={1},
	year={2018}
}

@article{wu2024deep,
	title={Deep multimodal learning with missing modality: A survey},
	author={Wu, Renjie and Wang, Hu and Chen, Hsiang-Ting and Carneiro, Gustavo},
	journal={arXiv preprint arXiv:2409.07825},
	year={2024}
}

@article{reza2024robust,
	title={Robust multimodal learning with missing modalities via parameter-efficient adaptation},
	author={Reza, Md Kaykobad and Prater-Bennette, Ashley and Asif, M Salman},
	journal={IEEE TPAMI},
	year={2024},
	publisher={IEEE}
}

@article{guo2024multimodal,
	title={Multimodal prompt learning with missing modalities for sentiment analysis and emotion recognition},
	author={Guo, Zirun and Jin, Tao and Zhao, Zhou},
	journal={arXiv preprint arXiv:2407.05374},
	year={2024}
}

@inproceedings{dai2025unbiased,
	title={Unbiased missing-modality multimodal learning},
	author={Dai, Ruiting and Li, Chenxi and Yan, Yandong and Mo, Lisi and Qin, Ke and He, Tao},
	booktitle={ICCV},
	pages={24507--24517},
	year={2025}
}

@article{liu2024modality,
	title={Modality translation-based multimodal sentiment analysis under uncertain missing modalities},
	author={Liu, Zhizhong and Zhou, Bin and Chu, Dianhui and Sun, Yuhang and Meng, Lingqiang},
	journal={Information Fusion},
	volume={101},
	pages={101973},
	year={2024},
	publisher={Elsevier}
}

@article{li2024toward,
	title={Toward robust incomplete multimodal sentiment analysis via hierarchical representation learning},
	author={Li, Mingcheng and Yang, Dingkang and Liu, Yang and Wang, Shunli and Chen, Jiawei and Wang, Shuaibing and Wei, Jinjie and Jiang, Yue and Xu, Qingyao and Hou, Xiaolu and others},
	journal={NIPS},
	volume={37},
	pages={28515--28536},
	year={2024}
}

@article{xie2024trustworthy,
	title={Trustworthy multimodal fusion for sentiment analysis in ordinal sentiment space},
	author={Xie, Zhuyang and Yang, Yan and Wang, Jie and Liu, Xiaorong and Li, Xiaofan},
	journal={IEEE TCSVT},
	volume={34},
	number={8},
	pages={7657--7670},
	year={2024},
	publisher={IEEE}
}

@inproceedings{zeng2022mitigating,
	title={Mitigating inconsistencies in multimodal sentiment analysis under uncertain missing modalities},
	author={Zeng, Jiandian and Zhou, Jiantao and Liu, Tianyi},
	booktitle={EMNLP},
	pages={2924--2934},
	year={2022}
}

@inproceedings{mao2023robust,
	title={Robust-MSA: Understanding the impact of modality noise on multimodal sentiment analysis},
	author={Mao, Huisheng and Zhang, Baozheng and Xu, Hua and Yuan, Ziqi and Liu, Yihe},
	booktitle={AAAI},
	volume={37},
	number={13},
	pages={16458--16460},
	year={2023}
}

@article{yu2024spikemo,
	title={SpikEmo: Enhancing Emotion Recognition With Spiking Temporal Dynamics in Conversations},
	author={Yu, Xiaomin and Wang, Feiyang and Qiao, Ziyue},
	journal={arXiv preprint arXiv:2411.13917},
	year={2024}
}

@article{yu2026modality,
	title={Modality gap-driven subspace alignment training paradigm for multimodal large language models},
	author={Yu, Xiaomin and Xin, Yi and Zhang, Yuhui and Zhang, Wenjie and Liu, Chonghan and Zhao, Hanzhen and Liu, Chen and Hu, Xiaoxing and Qiao, Ziyue and Tang, Hao and others},
	journal={arXiv preprint arXiv:2602.07026},
	year={2026}
}

@article{yu2026anisotropic,
	title={Anisotropic Modality Align},
	author={Yu, Xiaomin and Li, Yijiang and Zhang, Yuhui and Zhao, Hanzhen and Yang, Yue and Tang, Hao and Song, Yue and Hu, Xiaobin and Qin, Chengwei and Yan, Shuicheng and others},
	journal={arXiv preprint arXiv:2605.07825},
	year={2026}
}
	
\end{document}